\documentclass[10pt,twocolumn,letterpaper]{article}
\usepackage{colortbl} 

\usepackage[pagenumbers]{cvpr} 
\usepackage{tabularx}

\usepackage[dvipsnames]{xcolor}

\usepackage{etoolbox}

\definecolor{cvprblue}{rgb}{0.21,0.49,0.74}
\usepackage[pagebackref,breaklinks,colorlinks,citecolor=cvprblue]{hyperref}

\def\paperID{2252} 
\def\confName{CVPR}
\def\confYear{2024}

\newcommand{\mbtheta}{\mathbf{\theta}}
\newcommand{\mbp}{\mathbf{p}}
\newcommand{\mbx}{\mathbf{x}}
\newcommand{\mbv}{\mathbf{v}}
\newcommand{\mbt}{\mathbf{t}}
\newcommand{\mbc}{\mathbf{c}}
\newcommand{\tvgg}{\text{VGG}}

\definecolor{best_color}{HTML}{FB9999}
\definecolor{better_color}{HTML}{FDCC99}
\definecolor{good_color}{HTML}{FFF8AE}

\title{SplatArmor: Articulated Gaussian splatting for animatable humans from monocular RGB videos }

\author{
Rohit Jena\textsuperscript{1}\thanks{Work done outside of Amazon} 
\and Ganesh Iyer\textsuperscript{2}
\and Siddharth Choudhary\textsuperscript{2}  
\and Brandon M. Smith\textsuperscript{2} 
\and Pratik Chaudhari\textsuperscript{1}
\and James C. Gee\textsuperscript{1}  \\ 
{
\hspace{-6cm}\textsuperscript{1}University of Pennsylvania \hspace{1cm} \textsuperscript{2}Amazon.com, Inc} 
}

\definecolor{brandoncolor}{RGB}{77, 157, 224}
\definecolor{rohitcolor}{RGB}{225, 85, 84}
\definecolor{sidcolor}{RGB}{59, 178, 115}
\definecolor{ganeshcolor}{RGB}{150, 30, 194}

\newcolumntype{Y}{>{\centering\arraybackslash}X}

\makeatletter
\apptocmd{\maketitle}{{\myfigure{}\par}}{}{}
\makeatother

\begin{document}

\newcommand\myfigure{%
}

\maketitle

\begin{abstract}
We propose SplatArmor, a novel approach for recovering detailed and animatable human models by `armoring' a parameterized body model with 3D Gaussians.
Our approach represents the human as a set of 3D Gaussians within a canonical space, whose articulation is defined by extending the skinning of the underlying SMPL geometry to arbitrary locations in the canonical space. 
To account for pose-dependent effects, we introduce a SE(3) field, which allows us to capture both the location and anisotropy of the Gaussians.
Furthermore, we propose the use of a neural color field to provide color regularization and 3D supervision for the precise positioning of these Gaussians.
We show that Gaussian splatting provides an interesting alternative to neural rendering based methods by leverging a rasterization primitive without facing any of the non-differentiability and optimization challenges typically faced in such approaches.
The rasterization paradigms allows us to leverage forward skinning, and does not suffer from the ambiguities associated with inverse skinning and warping.
We show compelling results on the ZJU MoCap and {People Snapshot} 
datasets, which underscore the effectiveness of our method for controllable human synthesis.
\end{abstract}
    
\section{Introduction}
\label{sec:intro}

Our goal is to generate detailed, personalized, and animatable 3D human models, from monocular RGB videos.
This has many downstream applications, such as customized virtual reality avatars, teleconferencing, and realistic synthetic data generation.
Unlike marker-based 3D motion capture and body scanning systems, generating human avatars from video is inexpensive. 
Markerless human capture, whether from monocular or multi-view videos, offers a convenient and accessible means to achieve high-fidelity controllable 3D avatars of the human body.

Initial approaches to recover a human avatar from RGB videos relied on using an artist-defined mesh  topology with rigging and optimizing its geometry and texture.
Several approaches have been proposed to recover coarse shape and pose~\cite{Kanazawa18cvpr,Kocabas20cvpr,Sengupta20bmvc,Sengupta21iccv,Sengupta21cvpr,Sengupta21bmvc,Smith193dv,Yu21iccv,Yu22cvpr,Zhang21iccv,xiang2020monoclothcap}, jointly recover the shape, pose, and texture~\cite{alldieck:3dv2018:humanmono,alldieck:cvpr2018:peoplesnap}.
However, these methods are hard to optimize, and face difficulty in recovering geometry that does not conform to the topology of the underlying mesh.
Moreover, it is non-trivial to capture pose-dependent effects in the mesh.
Recently, human-specific neural rendering methods have demonstrated state-of-the-art results in controllable human synthesis~\cite{Saito19pifu,Saito20pifuhd,weng:cvpr2022:humannerf,chen:arxiv2021:animner,peng2021animatable,Alldieck22phorhum,xu2022sanerf,bhatnagar:eccv2020:ipnet,carranza2003free}.
These methods utilize neural representations of geometry (continuous density functions, SDFs), allowing for modelling substantial geometric deviations from standard shape models, accomodating different topologies, and effectively addressing pose-dependent effects.
Volumetric rendering is then employed using a raytracing approach to synthesize 2D renders of the subject.
Recently, Gaussian Splatting~\cite{kerbl3Dgaussians} has been shown to be an effective alternative representation to NeRFs for static and dynamic scenes.

\begin{figure*}[h!]
   \centering 
   \includegraphics[width=0.95\textwidth]{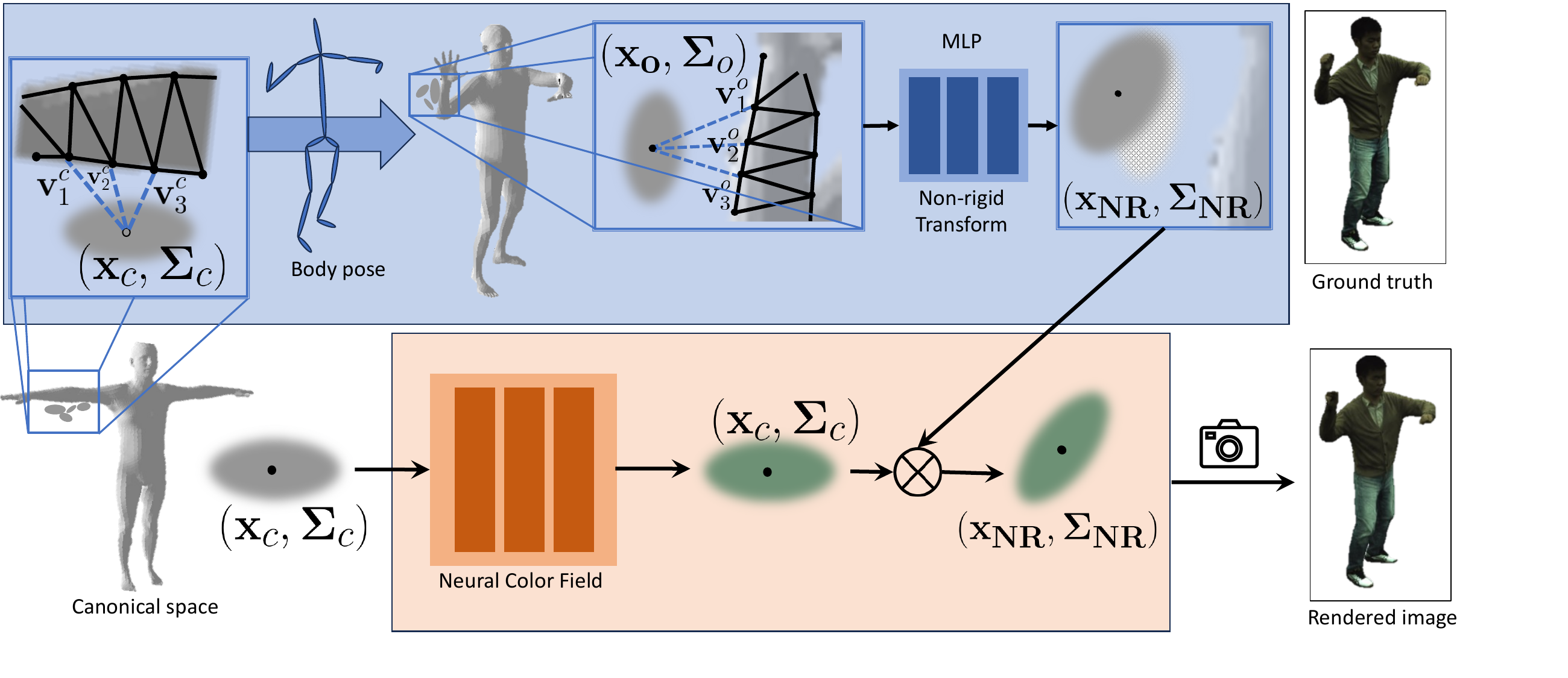}
   \caption{\textbf{Overview of our approach}. SplatArmor is defined as an SMPL mesh and a set of Gaussians in the canonical space.
   The transform and color modules are shown in {\textcolor{blue}{blue}} and {\textcolor{orange}{orange}} panels.
   For a Gaussian denoted by $(\mbx_c, \mathbf{\Sigma}_c)$, we find the k-nearest neighbors in the mesh (denoted as $\mbv_1^c, \mbv_2^c \ldots$). Given a body pose, the \textit{uncolored} Gaussians are moved according to the blend weights defined in Eq.~\ref{eq:lbs-p}. To capture pose dependent effects, a non-rigid transform MLP is added, determining the final Gaussian $(\mbx_{\text{NR}}, \mathbf{\Sigma}_{\text{NR}})$ in the observation space.
   This Gaussian is then \textit{colored} by querying a color field MLP using the canonical coordinate to provide the final observed Gaussian.
   }
   \label{fig:overall}
\end{figure*}

In this paper, we explore Gaussian Splatting to recover detailed and animatable human model from RGB videos.
Such an approach has several benefits. 
First, Gaussian Splatting utilizes rasterization, which is much faster than the raytracing approach used in NeRFs. 
Second, most NeRF-for-human methods perform inverse skinning for canonicalization, which can have ambiguous multiple solutions~\cite{Chen21SNARF,xu2022sanerf,zheng2022avatar,Su21anerf}.
This is because the points on the rays reside in the observation space.
In contrast, our approach utilizes Gaussian primitives within the canonical space, which are subsequently mapped to the observation space via forward skinning.
Forward skinning method avoids the correspondence ambiguities that are present in inverse skinning.
Third, Gaussians are also not topologically constrained unlike a mesh, and can therefore inherit the topology and geometry of the subject from data.
The use of 3D Gaussians employs a rasterization paradigm, making it fast but avoiding the challenges associated with optimizing other rasterization primitives in terms of differentiability~\cite{liu2019softras,kato2018renderer,loper:eccv2014:opendr}.
Overall, this formulation presents an excellent solution for achieving realistic textures of a human from monocular RGB video, leveraging an underlying `coarse' geometric model to `anchor' or `armor' the Gaussians around the model.

\section{Related Work}
\label{sec:related}

\textbf{Neural rendering for human recovery}:
Neural representations~\cite{Mildenhall20nerf,Xie22neuralfields} have led to compelling results for recovering human geometry and texture.
Recent methods such as PiFu~\cite{Saito19pifu}, PiFuHD~\cite{Saito20pifuhd}, and PHORUM~\cite{Alldieck22phorhum} learn an implicit representation based on pixel-aligned image features.
Other methods combine both implicit and explicit representations of geometry~\cite{Corona22lvd,bhatnagar:eccv2020:ipnet,Zheng20pamir,Cao22jiff,Xiu22icon} to represent clothed people.
However, these methods regress the geometry from a single (or few images) and do not utilize test-time optimization to correct inaccuracies/ambiguities in the predicted representation.
Recently, NeRF-style training has been extended to represent human avatars~\cite{peng:cvpr2021:neuralbod,Su21anerf,chen:arxiv2021:animner,weng:cvpr2022:humannerf,jiang:eccv2022:neuman,Weng20vid2actor,Liu21neuralactor,wang:eccv2022:arah,Jiang22selfrecon,feng2022capturing,Te22neuralcapture,dong:cvpr2022:pina}.
Neuralbody~\cite{peng:cvpr2021:neuralbod} uses structured latent codes based on the posed SMPL~\cite{SMPL:2015} mesh to produce a per-frame NeRF.
Peng \etal~\cite{peng2021animatable} use latent codes to produce a per-frame inverse blend skinning field.
However, per-frame latent codes overfit to the training frames leading to poor novel pose synthesis.
HumanNeRF~\cite{weng:cvpr2022:humannerf} instead learns a forward skinning weight field to avoid this overfitting, and derive the inverse skinning weights.
A-NeRF~\cite{Su21anerf} uses an articulated skeleton pose model and uses a skeleton-relative encoding with relative coordinates and directions to feed into a NeRF.
NARF~\cite{noguchi2021narf} uses a similar formulation by representing a global coordinate into local coordinates relative to each bone, and then querying a part-specific NeRF.
SelfRecon~\cite{Jiang22selfrecon} learns an SDF with forward skinning weights and uses a derived mesh to approximate intersection points with the SDF.
Other NeRF approaches~\cite{Liu21neuralactor,chen:arxiv2021:animner,xu2022sanerf} use a SMPL mesh to anchor a point from the observation space back into the canonical space.
NeuralActor~\cite{Liu21neuralactor} also uses a texture rendering module to resolve uncertainty from the ambiguities in inverse skinning and mapping from skeletal pose to dynamic effects.

\textbf{Dynamic Gaussian Splatting}:
Gaussian Splatting~\cite{kerbl3Dgaussians} represents a static scene using 3D Gaussians that preserve the properties of volumetric rendering without using expensive raytracing.
Gaussian splatting has been extended to dynamic scenes by learning a temporal dynamics that govern the movement of the Gaussians.
Luiten \etal~\cite{luiten2023dynamic} represent dynamic scenes by an analysis-by-synthesis framework, by allowing Gaussians to move and rotate freely over time while enforcing persistence of color, opacity and size.
The free form movement (with local rigidity constraints) allows persistent synthesis over time when the Gaussians represent the 3D scene fully.
Yang \etal~\cite{yang2023deformable} extend static Gaussians by learning a time-dependent deformation field that transports the Gaussians into the observed frame space.
Wu \etal~\cite{wu20234d} use multi-resolution HexPlanes to compute spacetime voxel features which are extracted from the centers of 3D Gaussians.
These features go through a tiny MLP that deforms the position, rotation and scale of the gaussians, which are then used to render the frame.
Xu \etal~\cite{xu20234k4d} use a similar idea to use K-planes to represent a 4D feature vector, and utilize a differentiable depth peeling algorithm for faster training.
These methods primarily focus on novel view rendering and do not provide grounding with an underlying geometry/animatable model.

\textbf{Concurrent Work}: Zielonka \etal~\cite{Zielonka2023Drivable3D} propose layered drivable 3D Gaussians for human recovery. 
They focus on dense multi-view setups (200 synchronized cameras) and embed the Gaussians in tetrahedral cages.
This method is trained on 12000 images, in contrast to our method not requiring more than 110 images.
Moreover, the number of optimizable Gaussians is fixed, which may not adapt to varying texture level in different subjects.
Instead, our method starts with a low number of Gaussians and uses adaptive density control depending on reconstruction quality.
Our method also focuses on reconstruction from monocular videos, which is a more accessible and natural way to capture in-the-wild human subjects.

\section{Method}
\label{sec:method}
\textbf{Problem Formulation}: Given a set of $N$ images $\{\mathcal{I}_t\}_{t=1}^N$ with associated foreground masks $\{\mathcal{F}_t\}_{t=1}^N$ and initial SMPL parameters $\beta, \{\theta_t\}_{t=1}^N$, our approach recovers the SMPL shape parameter $\beta^*$, per-vertex deformation $\mathbf{D}$, per-frame body poses and camera extrinsics $\{\theta_t^*, {E}_t^*\}$ and a set of Gaussians $\mathcal{S} = \{\mbx_i, \mathbf{s}_i, \mathbf{q}_i, \mathbf{\alpha}_i, C_i\}_{i=1}^{N}$.
The 3D Gaussians reside in a canonical space, and are transformed into the observation space to render the subject across frames.

An overview of our approach is shown in Fig. ~\ref{fig:overall}. 
The Gaussians in the canonical space are articulated by extending the blend skinning of the underlying SMPL mesh to arbitrary points in 3D space (Sec~\ref{sec:blendskinning}), and capturing additional pose-dependent non-rigid deformation (Sec~\ref{sec:posedep}).
Unlike existing methods, we do not optimize per-Gaussian colors, but instead propose a novel neural color field (Sec~\ref{sec:colorfield}) to implicitly regularize the color of nearby Gaussians.
The neural color field also provides 3D supervision to the Gaussian means.
Finally, we describe the details for optimizing a SplatArmor (Sec~\ref{sec:optim}) and an elegant initialization scheme for both the Gaussians and the color field.

\subsection{Extending blend skinning for 3D Gaussians}
\label{sec:blendskinning}
Typical NeRF methods articulate points on the canonical space by either using a rigged skeleton to define blend weights~\cite{weng:cvpr2022:humannerf,noguchi2021narf,Su21anerf,peng2021animatable} 
or using an underlying SMPL mesh to define, and optionally finetune the blend weights~\cite{chen:arxiv2021:animner,Jiang22selfrecon,Liu21neuralactor}.
We adopt the latter approach and use the SMPL mesh to define the blend weights for the entire space, since we can leverage an initialized SMPL template that matches the coarse geometry of the human.

Given a target frame or pose, rendering humans with NeRFs is typically done by sampling points on rays in the observation space, and inverting them into the canonical space.
However, as noted in existing works~\cite{Chen22gdna,Chen21SNARF,xu2022sanerf}, inverse skinning is pose dependent and may lead to overfitting or multiple solutions for novel poses.
In contrast, the 3D Gaussians lie on the canonical space, and they can be transported to the desired locations by extending the forward skinning algorithm defined by the SMPL model.

For an SMPL model with template vertices $\mathbf{T} = \{\mathbf{v^{c}_i}\}_{i=1}^{|\mathcal{V}|}$ residing in the canonical space, and body pose ${\theta}$, the posed vertices in the observed space are defined as the linear blend skinning (LBS) equation:
\begin{equation}
    \mathbf{v^{o}_i} = 
    \sum_{j=1}^{|\mathcal{J}|} \omega_{i,j} \left( \mathcal{G}_{j}({\mbtheta})
    \mathbf{v^{c}_i} + \mathbf{t}_{j}(\theta) \right) = \mathcal{M}_{i}(\mbtheta) \mathbf{v^c_i} + \mathbf{t}(\theta)
\end{equation}
where $\mathcal{G}_j(\mbtheta), \mathbf{t}(\mbtheta)$ defines the rigid motion of joint $j$ under joint rotations defined by $\mbtheta$ and $\omega_{i,j}$ are the blend weights of vertex $i$ with joint $j$.
This equation is only defined on the vertices of the template mesh.
To extend this idea for any general point $\mbx$, we use a similar formulation as~\cite{Zheng20pamir,chen:arxiv2021:animner} and define the forward skinning for an arbitrary point $\mbx$ in the canonical space as:

\begin{align}
    \mbx^o &= \sum_{i \in \mathcal{N}(\mbx)} \tau_i(\mbx) (\mathcal{M}_i(\mbtheta) \mbx + \mathbf{t}(\theta)) 
    \label{eq:tau_i}  \\
    &= \mathcal{M}{(\mbtheta, \mbx)}\mbx + \mathbf{t}(\theta, \mbx)
    \label{eq:lbs-p} 
\end{align}
where $\mathcal{N}(\mbx)$ denote the k-nearest neighbor SMPL vertices of $\mbx$, and the weights $\tau_i$ are defined as
\begin{align}
    \hat{\tau}_i(\mbx) &= \exp{\left(-\frac{\| \mathbf{v^o_i} - \mbx \|\| \mathbf{\omega}_i - \mathbf{\hat{\omega}} \|}{2\sigma^2}\right)} \\
    \tau &= {\sum_{i \in \mathcal{N}(\mbx)}\hat{\tau}_i} \quad \quad \text{and} \quad\quad \tau_i = {\hat{\tau_i}}/\tau
\end{align}
where $\mathbf{\hat{\omega}}$ is the blend weight vector of vertex $i$ and $\mathbf{\hat{\omega}}$ is the blend weight vector of the nearest neighbor of $\mbp$.
We drop the arguments of $\mathcal{M}$ 
for brevity, wherever it is clear from context.
Note that $\tau_i$ in Equation~\ref{eq:tau_i} is independent of the pose $\mbtheta$, and therefore, Equation~\ref{eq:tau_i} defines pose-independent LBS weights for an arbitrary point $\mbx$.
Intuitively, this extends the notion of blend skinning - the vertices on the mesh are rigidly controlled by the joints, and the points around the surface are rigidly controlled by the nearest set of vertices, which act as `virtual joints'.
Since the point $\mbx$ moves according to the rigid motion $\mathcal{M}(\mbtheta, \mbx)$, a Gaussian located at $\mbx$ with covariance $\Sigma$ in canonical space 
has an observation covariance
$\Sigma^o = \mathcal{M} \Sigma \mathcal{M}^T $.
This models the rigid motion of any Gaussian in the canonical space for arbitrary pose $\mbtheta$.

\subsection{Pose-dependent non-rigid deformation}
\label{sec:posedep}

Non-rigid motion in dynamic NeRFs is typically modeled as an offset field $\mathbf{\Delta x}$ conditioned on the body pose $\theta$~\cite{Liu21neuralactor,weng:cvpr2022:humannerf,Jiang22selfrecon,peng2021animatable}.
Owing to the anisotropic nature of the Gaussians, we model the pose-dependent motion of the Gaussian using a rigid transform instead of an offset.
\begin{equation}
    \mathcal{A}_{\text{NR}}(\mathbf{x}), \mathbf{t}_{\text{NR}}(\mathbf{x}) 
    = \text{MLP}_{\phi_{\text{NR}}}(\gamma(\mathbf{x}); \gamma_p(\theta))
    \label{eq:nonrigid-mlp}
\end{equation}
where $\gamma$ is the standard positional encoding, and $\gamma_p$ is the pose feature used in SMPL (\ie, $\gamma_p(\theta) = \exp(\theta) - I$).
This is because the axis-aligned representation of the body pose $\theta$ has the same low-frequency bias as spatial coordinates $\mbx$.
Note that for a Gaussian at canonical coordinate $\mbx$, the final observed location is given by $\mathcal{A}_{\text{NR}}(\mbx)(\mathcal{M}(\theta, \mbx) \mbx + \mbt(\theta, \mbx)) + \mathbf{t}_{\text{NR}}(\mbx)$
and the observed covariance matrix is $$\Sigma^o = \mathcal{A}\mathcal{M}\Sigma\mathcal{M}^T\mathcal{A}^T$$

\subsection{Neural Color Field}
\label{sec:colorfield}
A straightforward approach to represent a set of Gaussians is to assign a color (or spherical harmonics coefficients) for each Gaussian independently~\cite{luiten2023dynamic,yang2023deformable,kerbl3Dgaussians}.
However, for dynamic scenes where the body pose, pose dependent effects, and canonical space is optimized jointly, the per-Gaussian colors tend to overfit to the training frames, resulting in spurious texture artifacts. 
This leads to poor rendering performance on test frames (Sec. ~\ref{sec:ablation}).
An initial strategy to mitigate this behavior is to 
apply ad-hoc regularization on the colors of nearby Gaussians.
However, this will require calculating nearest neighbors for each Gaussian, whose time complexity is quadratic in the number of Gaussians.
We propose an alternate strategy to model the color of the Gaussian as a neural color field represented by an MLP
\begin{equation}
    C(\mbx) = \text{MLP}_{\phi_C}(\gamma(\mbx))
    \label{eq:color-mlp}
\end{equation}
This representation has two advantages.
First, the MLP provides implicit regularization 
to the color as a function of $\mbx$~\cite{deepimageprior}. 
Second, the learned color field provides an additional 3D supervision signal to the locations of the Gaussians.
Consider the case where a Gaussian with optimizable location $\mbx$ and color $\mbc$ is used to render an image.
The gradient $\nabla_\mbx\mathcal{L}_{\text{Render}}$ is obtained from the Gaussian renderer.
When the color is instead obtained using a neural color field $\mbc = C(\mbx)$, the derivative of $\mbx$ is given by
\begin{equation}
    \frac{\partial \mathcal{L}}{\partial \mbx} = \nabla_\mbx\mathcal{L}_{\text{Render}} + \frac{\partial L}{\partial C(\mbx)} \frac{\partial C(\mbx)}{\partial \mbx}
    \label{eq:3dsup}
\end{equation}

The Jacobian $\frac{\partial C}{\partial \mbx}$ provides information about local color changes at $\mbx$.
The second term in Equation~\ref{eq:3dsup} projects the rate of change of color
obtained by the renderer
with the Jacobian to obtain 3D supervision on $\mbx$.
Note that this network is used only during training; at inference the Gaussians are fixed, therefore the colors can be queried only once and cached during inference.

\begin{figure*}
    \centering
    \includegraphics[width=0.92\linewidth]{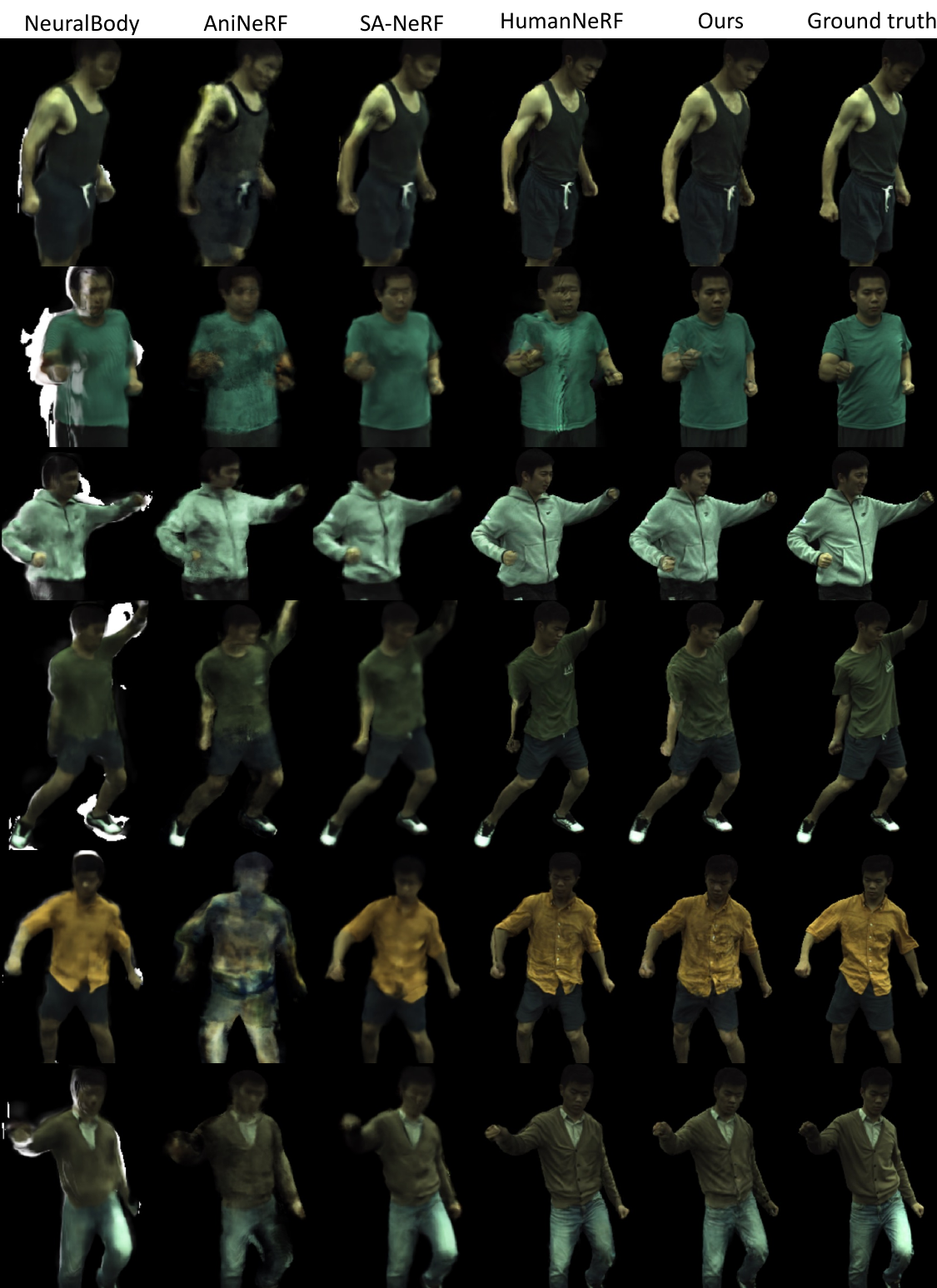}
    \caption{\textbf{Qualitative results on ZJU-MoCap dataset}. Best viewed when zoomed in.}
    \label{fig:zju-qual}
\end{figure*}

\begin{figure*}
    \centering
    \includegraphics[width=\textwidth]{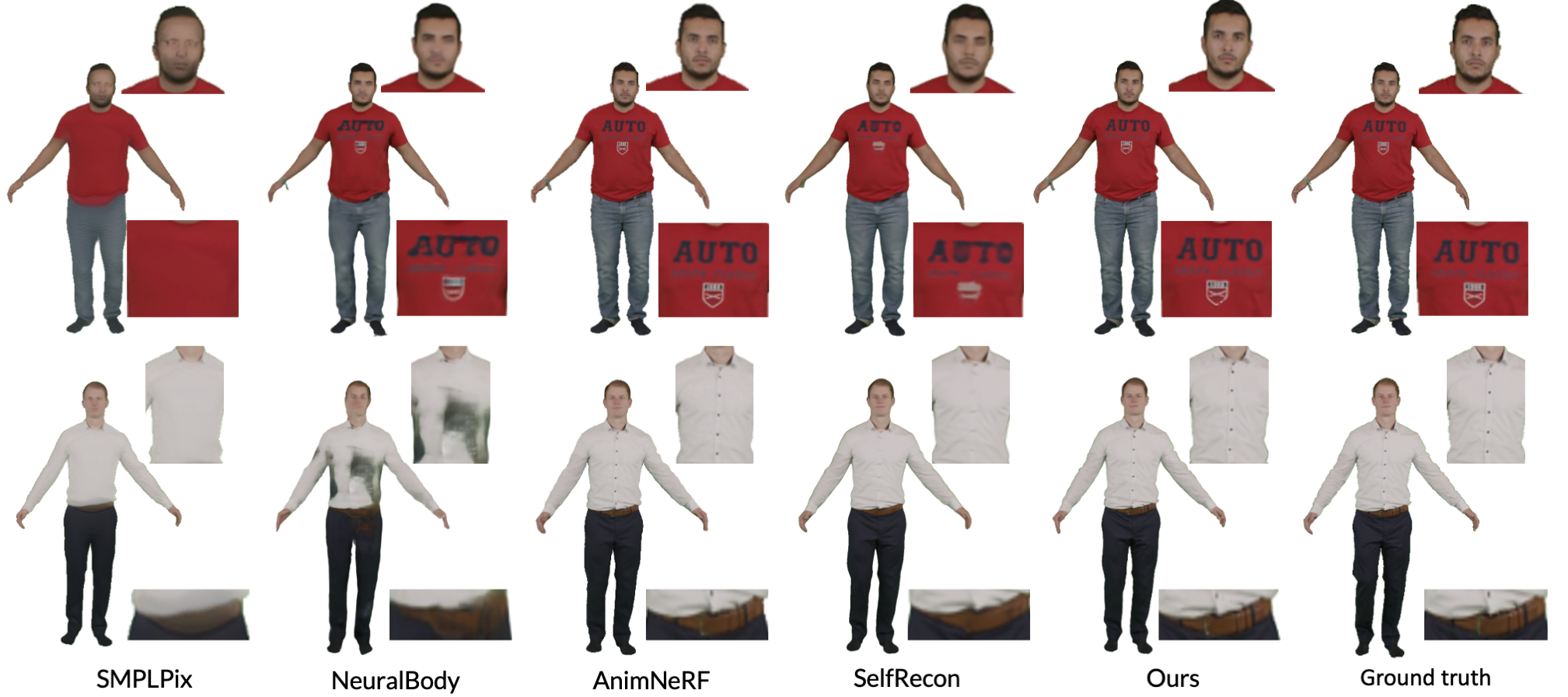}
    \caption{\textbf{Results on People Snapshot}. Our method performs competitively with state-of-the-art neural rendering approaches.}
    \label{fig:pplsnap-qual}
\end{figure*}

\section{Optimization details}
\label{sec:optim}
In this section, we describe the overall training strategy for optimizing 3D Gaussians.

\subsection{Initializing Gaussians and Neural Color Field}
For an explicit representation like Gaussian splatting, initializing the Gaussians has been shown to improve performance~\cite{kerbl3Dgaussians,luiten2023dynamic,wu20234d}.
The Neural Color Field can also provide noisy 3D supervision if initialized randomly, leading to slow convergence.
Moreover, the fidelity of Equation~\ref{eq:lbs-p} may reduce for points that are far from the surface of the mesh.
Therefore, a good initialization of the geometry and texture is crucial to our method.
We use the optimization strategy proposed in Jena \etal~\cite{jena2023mesh} to recover a coarse SMPL+D mesh, and a per-face color.
This step is relatively inexpensive, taking about 5-7 minutes.
We sample 20000 points from the surface of this coarsely optimized mesh, with the associated face color as the Gaussians.
The sampled location and color pairs are used to train the Neural Color Field using supervised learning.
Sec.~\ref{sec:nocf} analyzes the comparison of different initialization strategies. 

\subsection{Loss functions}
We employ an L1 loss similar to~\cite{kerbl3Dgaussians} to match the rendered images with the ground truth frames.
We note that pixelwise L1 loss does not provide robustness to slight misalignments in body pose due to its effective receptive field of 1 pixel.
Therefore, we also employ a perceptual loss using a VGG encoder~\cite{zhang2018unreasonable}.
We also employ a silhouette loss to avoid overfitting to the background.
To render a binary mask from Gaussians, we set the color of the Gaussians to be $rgb = [1,1,1]$, and render this new `mask image'.
We use the Dice score as a mask loss. Our final loss is therefore $\mathcal{L} = \mathcal{L}_1 + \lambda_{\tvgg}\mathcal{L}_{\tvgg} + \lambda_{dice}\mathcal{L}_{dice}$.

\subsection{Training}
For a sampled frame $i$, we use the learnable parameters $\beta$, $\theta_i$ to compute the per-vertex transform $\mathcal{M}(\theta_i)$, which is used to transform the Gaussians into the observation space (Eq.~\ref{eq:lbs-p}). 
The Gaussian centers are also used to obtain the colors and non-rigid transformation parameters from the MLPs (Eq.~\ref{eq:nonrigid-mlp},~\ref{eq:color-mlp}).
This adds the pose-dependent deformation and assigns the color, which is used by the renderer to produce an RGB and mask image.
The loss functions described above are used to optimize the MLPs and free parameters $\beta, \theta_i, E_i$.
We train our method for 500 epochs.
More implementation details can be found in Appendix.

\begin{figure}
    \centering
    \includegraphics[width=\linewidth]{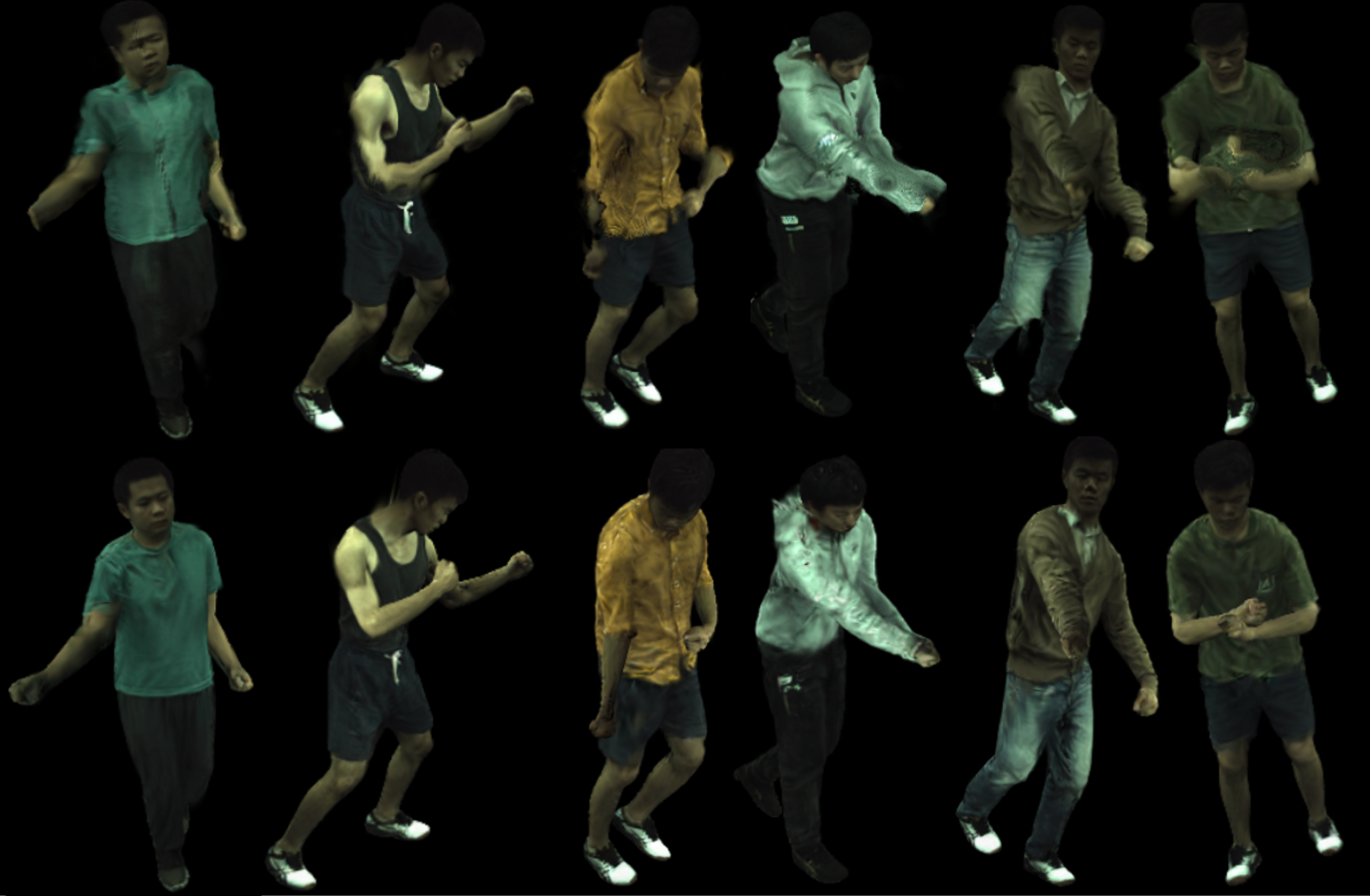}
    \caption{Results on unseen poses. HumanNeRF (top row) distorts the avatar severely, especially near the arms. Our method (bottom row) preserves the shape and fidelity of the rendering.}
    \label{fig:unseen-poses}
\end{figure}

\section{Evaluation}
\label{sec:evaluation}

\textbf{Datasets}:
We evaluate our method on the {People Snapshot}~\cite{alldieck:cvpr2018:peoplesnap} and ZJU-MoCap~\cite{peng:cvpr2021:neuralbod} datasets.
For people-snapshot, we select 4 subjects (male-3-casual, male-4-casual, female-3-casual, female-4-casual). 
We use the first 456 frames for training and the rest of the frames for validation.
For ZJU MoCap dataset, we consider 6 subjects (377,386,387,392,393,394) as in~\cite{weng:cvpr2022:humannerf}.
These subjects have loose clothing with wrinkles and large deformations, and have significantly harder poses. 
We use the first 450 frames in `camera 1' for training, and the rest of the frames in cameras 1,7,13,19 for novel view synthesis.

\begin{table*}[htbp]
\centering
\resizebox{\textwidth}{!}{%
\begin{tabular}{|c|c|c|c|c|c|c|c|c|c|c|c|c|}
\hline
& \multicolumn{2}{c|}{\textbf{Subject 377}} & \multicolumn{2}{c|}{\textbf{Subject 386}} & \multicolumn{2}{c|}{\textbf{Subject 387}} & \multicolumn{2}{c|}{\textbf{Subject 392}} & \multicolumn{2}{c|}{\textbf{Subject 393}} & \multicolumn{2}{c|}{\textbf{Subject 394}}  \\ \hline
& \textbf{PSNR}$\uparrow$ & \textbf{LPIPS*}$\downarrow$
& \textbf{PSNR}$\uparrow$ & \textbf{LPIPS*}$\downarrow$
& \textbf{PSNR}$\uparrow$ & \textbf{LPIPS*}$\downarrow$
& \textbf{PSNR}$\uparrow$ & \textbf{LPIPS*}$\downarrow$
& \textbf{PSNR}$\uparrow$ & \textbf{LPIPS*}$\downarrow$
& \textbf{PSNR}$\uparrow$ & \textbf{LPIPS*}$\downarrow$ \\ \hline \hline
\textbf{SMPLPix~\cite{smplpix}} & 27.00 & 90.74 & 30.38 & 97.91 & 23.80 & 114.76 & 29.12 & 72.66 & 24.79 & 126.50 & 26.99 & 84.47 \\
\textbf{NeuralBody~\cite{peng:cvpr2021:neuralbod}} & 23.84 & 67.18 & 23.26 & 55.01 & 23.15 & 67.75 & 22.46 & 70.36 & 22.41 & 71.32 & 22.19 & 72.90 \\
\textbf{AniNeRF~\cite{peng2021animatable}} & 22.32 & 53.93 & 25.03 & 55.53 & 15.08 & 189.94 & 23.27 & 76.71 & 19.51 & 82.86 & 21.46 & 78.89 \\
\textbf{SA-NeRF~\cite{xu2022sanerf}} & \cellcolor{better_color}32.04 & \cellcolor{good_color}33.01 & \cellcolor{better_color}35.25 & \cellcolor{good_color}37.31 & \cellcolor{better_color}29.73 & \cellcolor{good_color}55.23 & \cellcolor{better_color}32.26 & \cellcolor{good_color}54.58 & \cellcolor{better_color}30.16 & \cellcolor{good_color}58.43 & \cellcolor{better_color}30.68 & \cellcolor{good_color}55.69 \\
\textbf{HumanNeRF~\cite{weng:cvpr2022:humannerf}} & \cellcolor{good_color}29.72 & \cellcolor{better_color}26.31 & \cellcolor{good_color}32.55 & \cellcolor{better_color}36.44 & \cellcolor{good_color}28.37 & \cellcolor{better_color}30.85 & \cellcolor{good_color}30.91 & \cellcolor{better_color}34.86 & \cellcolor{good_color}28.66 & \cellcolor{better_color}36.39 & \cellcolor{good_color}29.09 & \cellcolor{better_color}41.43 \\
\textbf{Ours} & \cellcolor{best_color}33.06 & \cellcolor{best_color}19.77 & \cellcolor{best_color}35.57 & \cellcolor{best_color}25.42 & \cellcolor{best_color}30.03 & \cellcolor{best_color}30.73 & \cellcolor{best_color}32.48 & \cellcolor{best_color}33.20 & \cellcolor{best_color}30.24 & \cellcolor{best_color}32.56 & \cellcolor{best_color}31.41 & \cellcolor{best_color}30.07 \\
\hline
\end{tabular} %
}
\caption{\textbf{Quantitative comparison on ZJU MoCap dataset}.  LPIPS$^*$ = LPIPSx1000. \colorbox{best_color}{\hspace{0.15cm} \vphantom{X}} = First, \colorbox{better_color}{\hspace{0.15cm} \vphantom{X}} = Second, \colorbox{good_color}{\hspace{0.15cm} \vphantom{X}} = Third.}
\label{tab:zju}
\end{table*}

\textbf{Baselines}:
We choose a variety of baselines for both datasets.
To mitigate the effect of instrumentation bias, we consider baselines which either provide trained models or recommended training configurations.
For people-snapshot, we consider SMPLPix~\cite{smplpix}, NeuralBody~\cite{peng:cvpr2021:neuralbod}, AnimNeRF~\cite{chen:arxiv2021:animner} and SelfRecon~\cite{Jiang22selfrecon} as state-of-the-art baselines.
For ZJU MoCap, we consider SMPLPix which uses deferred rendering, NeuralBody, Animatable NeRF (AniNeRF)~\cite{peng2021animatable}, Surface-Aligned NeRF (SA-NeRF)~\cite{xu2022sanerf}, and HumanNeRF~\cite{weng:cvpr2022:humannerf} which are state-of-the-art neural rendering methods for animatable humans.

\begin{table*}[htbp]
\centering
\resizebox{0.9\textwidth}{!}{%
\begin{tabular}{|c|c|c|c|c|c|c|c|c|}
\hline
& \multicolumn{2}{c|}{\textbf{male-3-casual}} & \multicolumn{2}{c|}{\textbf{male-4-casual}} & \multicolumn{2}{c|}{\textbf{female-3-casual}} & \multicolumn{2}{c|}{\textbf{female-4-casual}} \\
\hline
& \textbf{PSNR}$\uparrow$ & \textbf{LPIPS*}$\downarrow$ & \textbf{PSNR}$\uparrow$ & \textbf{LPIPS*}$\downarrow$ & \textbf{PSNR}$\uparrow$ & \textbf{LPIPS*}$\downarrow$ & \textbf{PSNR}$\uparrow$ & \textbf{LPIPS*}$\downarrow$ \\
\hline \hline
\textbf{SMPLPix~\cite{smplpix}} & 17.90 & 165.74 & 17.23 & 198.82 & 17.35 & 135.91 & 18.24 & 150.11 \\
\textbf{NeuralBody~\cite{peng:cvpr2021:neuralbod}} & 20.16 & 72.37 & 19.43 & \cellcolor{good_color}84.55 & 18.67 & \cellcolor{good_color}80.35 & 19.98 & 66.65 \\
\textbf{AnimNeRF~\cite{chen:arxiv2021:animner}} & \cellcolor{better_color}25.01 & \cellcolor{better_color}44.92 & \cellcolor{good_color}23.28 & 89.59 & \cellcolor{good_color}21.19 & 89.94 & \cellcolor{good_color}24.60 & \cellcolor{best_color}52.00 \\
\textbf{SelfRecon~\cite{Jiang22selfrecon}} & \cellcolor{good_color}24.91 & \cellcolor{good_color}61.33 & \cellcolor{better_color}25.66 & \cellcolor{best_color}65.82 & \cellcolor{better_color}24.82 & \cellcolor{best_color}68.14 & \cellcolor{better_color}25.23 & \cellcolor{good_color}64.35 \\
\textbf{Ours} & \cellcolor{best_color}27.08 & \cellcolor{best_color}43.91 & \cellcolor{best_color}25.67 & \cellcolor{better_color}81.92 & \cellcolor{best_color}25.76 & \cellcolor{better_color}79.90 & \cellcolor{best_color}26.81 & \cellcolor{better_color}64.26 \\
\hline
\end{tabular}
}
\caption{\textbf{Quantitative comparison on People Snapshot dataset.}}
\label{tab:people-snap}
\end{table*}

\subsection{Comparison}
\label{sec:compare}

Quantitative results on ZJU MoCap dataset is shown in Table ~\ref{tab:zju}.
We note that AnimatableNeRF and SA-NeRF render blank images when trained with images from a single camera.
Therefore, we use cameras 1,7,13,19 for training for these baselines.
Quantitatively, our method consistently outperforms several strong baselines, with a notable improvement in LPIPS.
This is also evident qualitatively in Fig.~\ref{fig:zju-qual} where our renders preserve details like face, wrinkles and loose clothing.
NeuralBody tends to learn the background, as visible by the white artifacts.
SA-NeRF has very similar PSNR values to our method, but Fig.~\ref{fig:zju-qual} shows that it produces blurry results, showing the bias of PSNR towards smooth results~\cite{zhang2018unreasonable}.
HumanNeRF has good perceptual quality, but occasionally produces extreme deformations (second row in Fig.~\ref{fig:zju-qual}) or pose misalignments.

Table~\ref{tab:people-snap} shows the results for People Snapshot dataset.
Our method performs very competitively with state-of-the-art baselines such as AnimNeRF and SelfRecon.
Fig.~\ref{fig:pplsnap-qual} shows that all methods recover the geometry well, but SelfRecon relies on the textures obtained from VideoAvatar, leading to blurry results.
Our method synthesizes these subtle details (buttons, logo, belt) with high perceptual quality.

\textbf{Qualitative comparison on unseen poses}: 
In both datasets, the pose distribution in the training and validation frames are very similar.
In contrast, an animatable avatar should produce high-fidelity rendering on unseen and arbitrary poses.
To this end, we use the AMASS dataset~\cite{AMASS} to animate the trained models on the ZJU MoCap subjects due to its complexity.
Specifically, we select five sequences - WalkDog, BoxLift, SwitchStance, Aita and Hamada.
These poses are truly unseen and test the generalization ability of the methods.
We show an initial qualitative comparison with HumanNeRF, the best performing baseline, in Fig.~\ref{fig:unseen-poses}.
HumanNeRF distorts the avatars drastically on these sequences (especially near the arms), 
showing the limitations of inverse skinning for neural rendering.
In contrast, our formulation leverages the deformation of the underlying SMPL model, and maintains its fidelity across different poses.
A more comprehensive comparison on unseen poses for all baselines is provided in Supplementary Video.


\subsection{Ablation studies}
\label{sec:ablation}

\begin{table}[h!]
    \centering
    \begin{tabular}{|c|c|c|} \hline
    & \textbf{PSNR}$\uparrow$ & \textbf{LPIPS*} $\downarrow$ \\ \hline \hline
    translation & 31.55 & 28.64 \\
    affine & 31.32 & 28.82 \\ \hline
    w/o MSB~\cite{jena2023mesh} & 31.51 & 28.07 \\
    w/o pretraining CF  & \cellcolor{good_color}31.68 & \cellcolor{good_color}27.33 \\
    w/o Color Field  & \cellcolor{better_color}31.85 & \cellcolor{better_color}26.90 \\
    Ours & \cellcolor{best_color}31.94 & \cellcolor{best_color}26.08 \\
    \hline
    \end{tabular}
    \caption{{\textbf{Ablations} on pose MLP, pretraining and color network on ZJU MoCap}. Results are averaged over all 6 sequences.}
    \label{tab:ablations}
\end{table}

\textbf{Effect of neural color field}: 
\label{sec:nocf}
We ablate our method using optimizable vectors for the Gaussian colors, \ie no regularization.
This model leads to `stray Gaussians' that remain hidden in the training frames without any supervision applied to their color, since they do not contribute to rendering.
During novel pose synthesis, some of these Gaussians become visible, leading to artifacts. 
The Neural Color Field implicitly determines their color using the continuity of the MLP.

\begin{figure}
    \centering
    \includegraphics[width=0.8\linewidth]{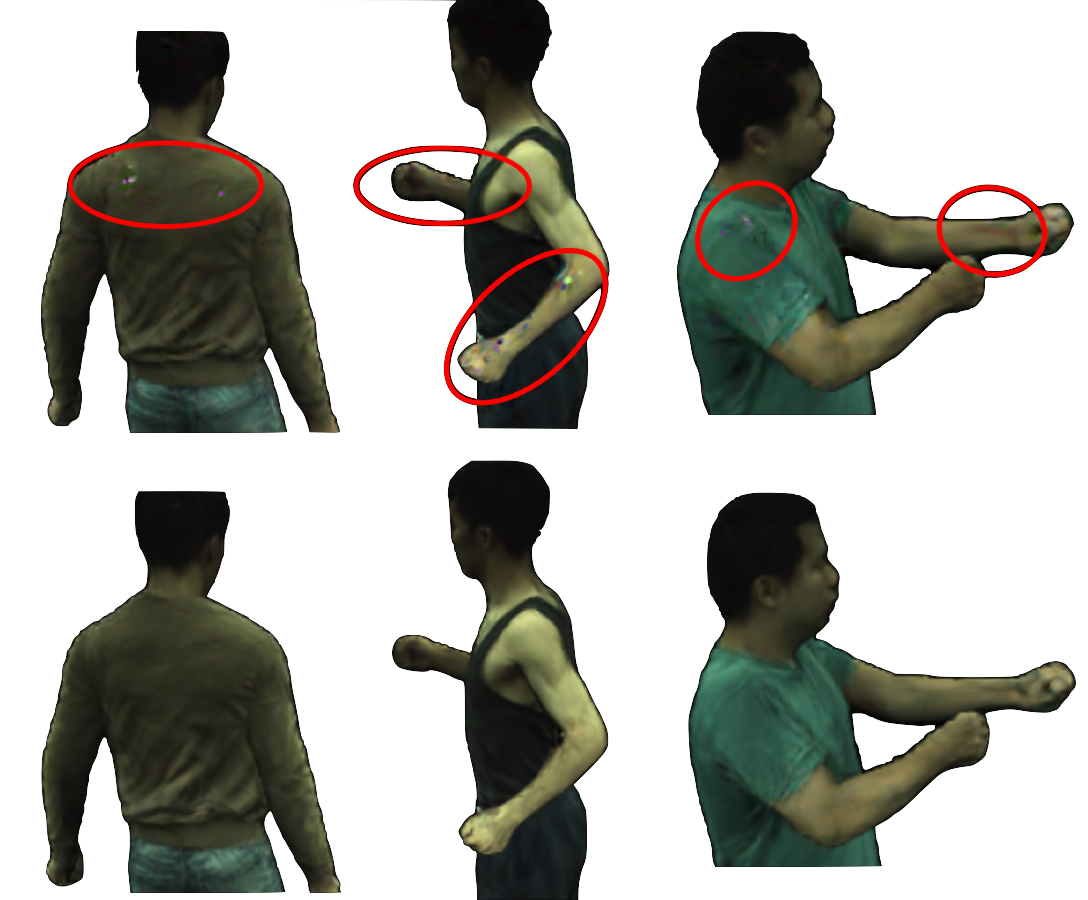}
    \caption{\textbf{Ablation on Color Field}. Top row denotes training with optimizable free parameters for colors similar to~\cite{kerbl3Dgaussians}. Bottom row are trained and rendered with Neural Color Field.}
    \label{fig:cf-ablation}
\end{figure}

\begin{figure}
    \centering
    \includegraphics[width=0.9\linewidth]{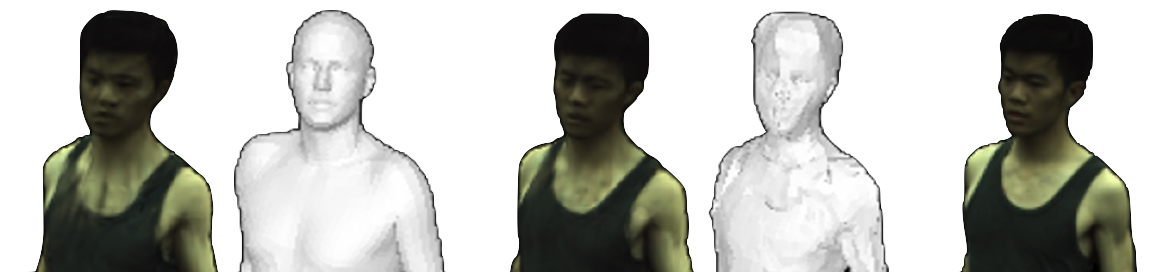}
    \includegraphics[width=0.9\linewidth]{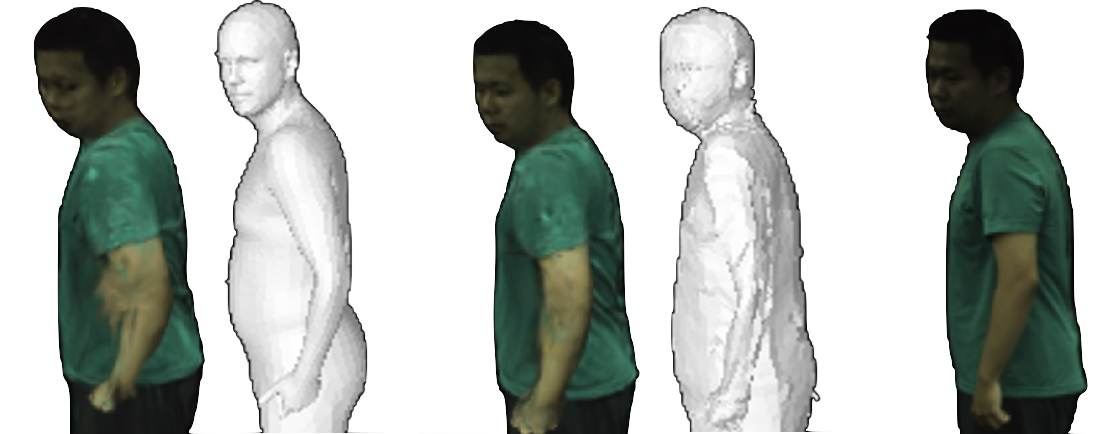}
    \caption{\textbf{Effect of pretraining.} Left shows renders from a model with no visual hull estimation. Middle shows renders from model initialized with ~\cite{jena2023mesh}. Right shows ground truth. }
    \label{fig:pretrain}
\end{figure}

\textbf{Effect of pre-training the visual hull and Neural Color Field}: 
Without a pre-training step, the underlying SMPL model does not reflect the actual geometry of the human, which has to be compensated by the non-rigid MLP.
On novel poses, the non-rigid MLP may fail to interpolate the movement correctly, leading to artifacts (Fig.~\ref{fig:pretrain}).

\begin{figure}
    \centering
    \includegraphics[width=\linewidth]{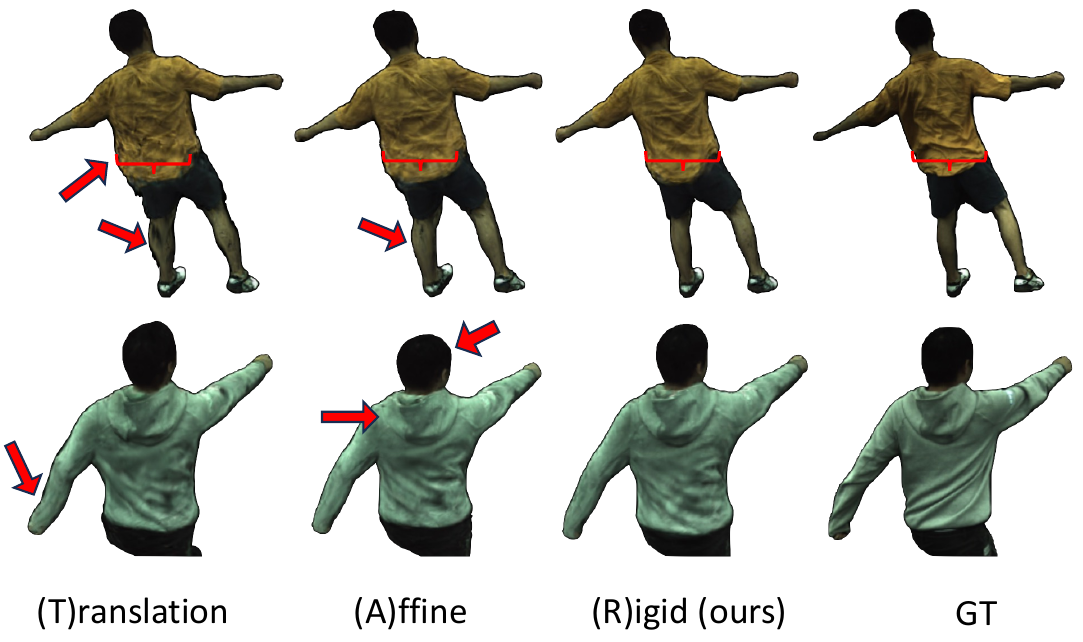}
    \caption{\textbf{Pose dependent transform}. Translation has artifacts due to inaccurate rotation of Gaussians, affine overcompensates, rigid provides balance between accuracy and overcompensation.}
    \label{fig:nonrigid}
\end{figure}

\textbf{Choice of pose-dependent MLP}: We consider 3 output choices for the pose-dependent MLP (Eq.~\ref{eq:nonrigid-mlp}): translation only (T), rigid (R), and full affine matrix (A). 
Qualitatively we observe that the (T) variant leads to noisy texture due to incorrect rotation of the Gaussians in novel poses. 
The (A) variant overcompensates the distortion in shape, evident by the squashed heads and wider torso in Fig.~\ref{fig:nonrigid}.
We use the (R) variant which provides the most accurate pose-dependent effects.

\subsection{Training and inference time}
ZJU MoCap is a challenging dataset, and a lot of training iterations are required to learn the pose dependent effects.
HumanNeRF is a strong baseline, but requires 3 days of training with around 48GB of GPU memory.
In contrast, our method can be trained in 7 hours with 9GB of GPU memory.
People Snapshot is a relatively easy dataset with minimal pose dependent effects.
For this dataset, our method converges in about 70 minutes, in contrast to AnimNeRF, which takes 15 hours.
Inference is real-time, unlike HumanNeRF and AnimNeRF which are effectively $<$ 0.2 FPS.
\section{Discussion}
\subsection{Limitations}
\begin{figure}[h!]
    \centering
    \includegraphics[width=0.8\linewidth]{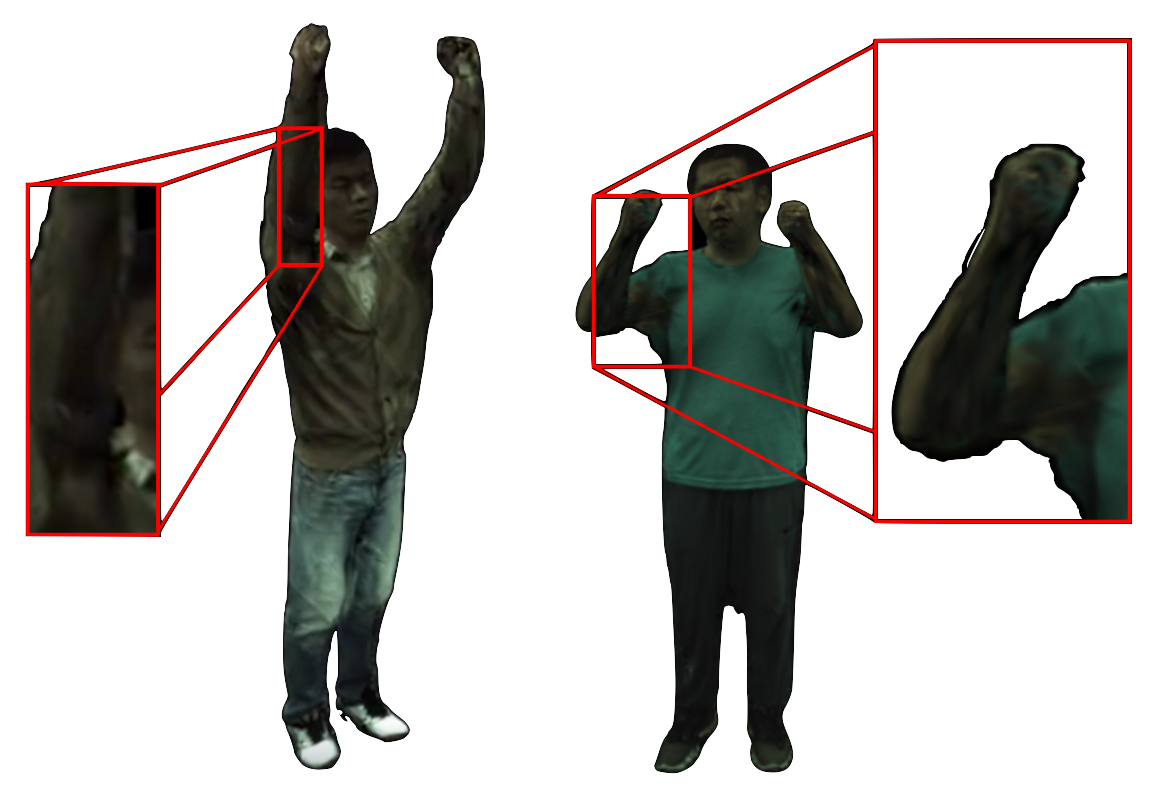}
    \caption{\textbf{Failure cases}. One of the failure modes we notice corresponds to the value of the Gaussian splatting in unseen areas.}
    \label{fig:failures}
\end{figure}

Jointly learning the color field, pose dependent dynamics, and coarse underlying geometry is a highly underconstrained problem.
Although we optimize the per-frame pose while training, bad initialization can lead to confounding signals to the misaligned Gaussians, leading to texture artifacts. 
Moreover, since we do not model view dependent colors similar to~\cite{weng:cvpr2022:humannerf}, we observe that unseen regions, or regions with self-shadows adopt a darker color, leading to inconsistent texture (Fig.~\ref{fig:failures}).
Moreover, we use linear blend skinning (LBS) to articulate the human. Although pose-dependent effects can compensate for the artifacts of LBS in the training poses, its generalization to unseen poses cannot be guaranteed.

\subsection{Conclusion}
We present SplatArmor, a method for producing state-of-the-art results for articulated humans.
We demonstrate very high fidelity results for novel view and pose generation by anchoring 3D Gaussians to a coarse mesh of the human. 
The pose-driven motion of the 3D Gaussians is modelled using a combination of extension of the LBS for SMPL, and an MLP for adding pose-dependent dynamics. 
A neural color field is proposed to regularize the colors of the Gaussians, and provide 3D supervision to the locations of these Gaussians.
An elegant pretraining scheme is proposed for high fidelity reconstruction.
The method requires very little compute, training, and inference time requirements compared to its NeRF counterparts, thus taking a solid step towards modelling humans and achieving photorealistic animatable human models.
An interesting direction for future work would be to account for unseen regions and inaccuracies and use generative models to inpaint these regions.
\clearpage
{
    \small
    \bibliographystyle{ieeenat_fullname}
    \bibliography{main}
}


\end{document}